# *Exposing ambiguities in a relation-extraction gold standard with crowdsourcing*

*Tong Shu Li, Benjamin M. Good\*, Andrew I. Su\**

*Department of Molecular and Experimental Medicine, The Scripps Research Institute, La Jolla, CA 92037*

**ABSTRACT**

Semantic relation extraction is one of the frontiers of biomedical natural language processing research. Gold standards are key tools for advancing this research. It is challenging to generate these standards because of the high cost of expert time and the difficulty in establishing agreement between annotators. We implemented and evaluated a microtask crowdsourcing approach that can produce a gold standard for extracting drug-disease relations. The aggregated crowd judgment agreed with expert annotations from a pre-existing corpus on 43 of 60 sentences tested. The levels of crowd agreement varied in a similar manner to the levels of agreement among the original expert annotators. This work reinforces the power of crowdsourcing in the process of assembling gold standards for relation extraction. Further, it highlights the importance of exposing the levels of agreement between human annotators, expert or crowd, in gold standard corpora as these are reproducible signals indicating ambiguities in the data or in the annotation guidelines.

## 1 INTRODUCTION

Structured networks of biological information are powerful tools for interpreting experimental results and guiding decision making. Computational methods for assembling such networks automatically from the scientific literature are extremely valuable. High-quality databases of manually annotated text provide the foundation for creating and evaluating such methods [1]. Such gold standards are generally created by teams of domain experts.

Assembling gold standard annotated corpora typically involves three broad phases [2]. First, the curation task is designed in the form of detailed annotation guidelines. Second, multiple annotators independently annotate a set of documents. Typically no more than two or three expert annotators annotate the same document because of the high costs involved. Third, measures of inter-annotator agreement are calculated. In cases where annotators disagreed, the annotation instructions could be revised, or consensus could be established through discussion. Inter-annotator agreement is a frequently used metric to measure the value of the annotated corpus.

Achieving high levels of agreement has proven very challenging, particularly for assembling standards for semantic relation extraction. Two important examples illustrate the difficulty. First, the team behind the SemRep information extraction system [3, 4] developed a corpus of semantic predications spanning more than 50 specific kinds of relationships extracted from sentences in PubMed abstracts [5]. Despite high levels of domain expertise and multiple rounds of refinement, they only achieved agreement levels in the range 0.415 to 0.536 (as measured by F-Measure between pairs of annotators) [5]. The second example is provided by the EU-ADR (European Union Adverse Drug Reaction) corpus [6]. In this case, annotators defined whether or not a relation existed between a given pair of entities in a sentence. (No indication of the semantic type of the relation was attempted.) Agreement between each annotator and the final standard was just 42.5% (with an increase to 74.7% if disagreements on the annotations of the concepts in a given relation were discarded). Both studies emphasized the large impact of different tools and workflows on the levels of agreement obtained.

These studies resulted in useful corpora that have, for example, been used to train and evaluate automated systems for relation extraction with good reported results (e.g., BeFree [7]) reported F measures in the range 0.80). However, we suggest that gold standard databases would be improved by reporting annotator consistency information at the level of individual annotations. For some annotations, there is no argument and all annotators agree, yet for others, multiple rounds of consensus building can fail. When evaluating or training a system based on these annotations it would be very useful to know the levels of agreement associated with each annotation. Does the automated extraction system fail on the same set of annotations that the experts had trouble agreeing on? Such information would make it possible to generate agreement-based scoring functions and to use this information to create different representations for training machine learning schemes.

In assembling this information, the community could and should take the simple step of including the annotator consistency information on a per-annotation basis. However (1) this information is not available for existing gold standards and (2) there are typically low numbers of annotators per gold standard annotation because the time of each expert annotator is costly. Because of (2), the per-annotation con-sistency signal is weaker and noisier than it could be with more annotators.

\* To whom correspondence should be addressed: asu@scripps.edu, bgood@scripps.edu

Here, we describe a solution to both of these challenges that employs the emerging practice of microtask crowdsourcing for generating annotated corpora [8-10]. Crowdsourcing provides an ideal method for identifying inconsistency driven by either inherent ambiguity in language or problems with the task design. Worker populations are heterogeneous; aggregating their individual decisions gives the ability to measure agreement at a fine level. We show that, for the EU-ADR relation extraction gold standard, our crowdsourcing approach can reproduce the standard with fair accuracy, identify the annotations for which the expert creators of the standard were inconsistent, and extend the standard by providing additional semantic information about the links between concepts.

## 2 METHODS

The objective of our crowdsourcing task was to reproduce and extend the relationship assignments for concept pairs identified in the EU-ADR corpus, focusing specifically on drug-disease relations. The EU-ADR corpus provides a collection of 300 PubMed abstracts annotated with relationships between drugs, genes, and diseases [6]. Each concept pair was linked with one of four relationship types: positive, speculative, negative (a statement about the lack of the relation), and false (a co-occurrence with no indicated relation). Each abstract was seen by three expert annotators, and only relationships independently mentioned by at least two annotators were included in the published EU-ADR data set. Relationships from only a single annotator were not included in the final version.

Our experimental dataset contained 60 sentences randomly chosen from the 244 sentences containing drug-disease interactions in the EU-ADR. We used the CrowdFlower crowdsourcing platform for our task. Workers were shown one sentence with the drug and disease highlighted in different colors, and asked to choose what the relationship was (Fig. 1). If the worker determined that there was either a positive or speculative relationship, then they were also asked to select the nature of the relationship from four choices (drug causes the disease, drug treats the disease, no more information, or additional information suggesting another relation type).

A total of 10 workers judged each sentence, and workers were paid 10 cents per sentence. In order to gain access to our tasks, each worker had to take a quiz of 10 test questions and achieve a minimum of 70% accuracy in order to work on the data. They were also tested with the same questions while working; if at any time their accuracy score fell below 70%, all of their work was rejected as inaccurate.

Individual judgments were aggregated to determine the crowd's response for each drug-disease relationship through a minor enhancement to majority voting. Instead of taking the choice with most votes, which in some cases led to ties, the choice with the top confidence score was chosen as the final answer. Confidence score is the sum of the accuracy scores of the workers who voted for that choice (as measured on the 10 test questions). The crowd agreement score for a particular task is the confidence score divided by the total confidence score for all choices. (In the case of all workers performing equally on the test questions, crowd agreement is simply percent agreement.)

## 3 RESULTS

The drug-disease relationship annotation task was completed within two hours of posting. A total of 168 workers took the quiz, of which 32 workers passed. The mean worker accuracy of those who passed the quiz was 85.54% (std dev 7.77%).

Correspondence with the EU-ADR is one method for assessing the accuracy of the crowd responses. If the crowd's choice with the highest agreement score matched the EU-ADR answer, crowd was defined to agree. When strict agreement was calculated using all four relationship types, the crowd achieved a 71.67% match with the EU-ADR (43/60). When speculative and positive are combined, as they were in the original evaluations of the EU-ADR corpus, agreement increases to 76.67% (46/60).

### 3.1 Crowd vs expert agreement

To test the premise that the crowd's disagreements would mirror expert disagreements, we compared our crowd data to the original raw annotator data for the EU-ADR (sent to us by the authors). We first analyzed the 20 relationships that achieved perfect consensus among three expert annotators. For these sentences, the crowd's answer agreed with the experts' answer in all 20 cases. We also found that the crowd in aggregate was confident in these assessments. As shown in Fig. 2, when three of three experts agree, the mean crowd agreement score is 92.08% (median 100%, std dev 14.90%). For the 40 relations where only two of three experts agree, the crowd's mean agreement score drops to 50.27%. While the variability in crowd responses increases substantially in the low-expert-agreement group, the difference was statistically significant (Student's unpaired t-test p = 1.151e-07 t = -6.0774). These data show that according to the EU-ADR guidelines, certain sentences are more difficult to judge, and this difficulty is shared by both experts and the crowd.

**Fig. 1.** Crowdsourcing interface for collecting semantic relation judgments.

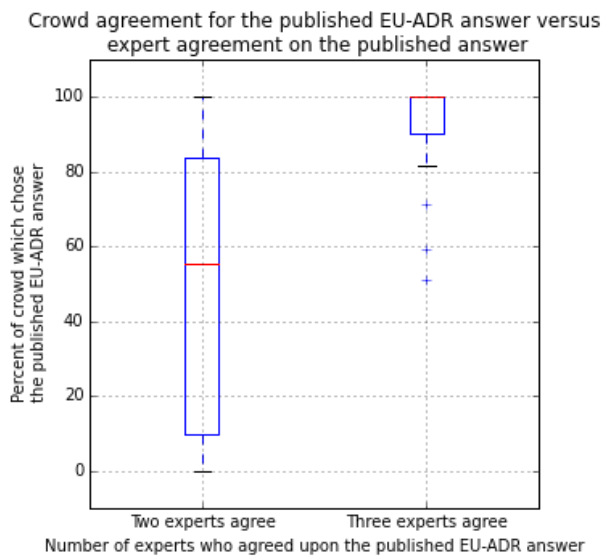

**Fig. 2.** Crowd agreement is significantly higher for relations where all three experts agreed on the published EU-ADR answer. Red line shows the median, box indicates the interquartile range (IQR), and the whiskers indicate 1.5-times IQR.

### 3.2 Disagreement analysis

Examination of sentences that were annotated differently by the experts and the crowd revealed several classes of disagreements. First, we identified cases that were truly ambiguous given the annotation guidelines. For example:

> "Dental surgery for these children has historically consisted of decreasing and/or discontinuing the oral **anticoagulant** and instituting heparin therapy prior to the planned dental procedure, which can result in **thromboembolism** and increased morbidity and mortality."

In this case, the sentence appears to positively link the combination of "anticoagulant" and "heparin" to "thromboembolism", but the annotation guidelines did not explicitly address how to handle the individual relationship between "anticoagulant" and "thromboembolism". EU-ADR annotators labeled this as "false", whereas the crowd chose "positive" (albeit with a very low crowd agreement score of 49.75%).

A second class of disagreements resulted from imprecise recognition of the concepts in the relationships. For example:

> "**Cardiovascular thromboembolic adverse effects** associated with **cyclooxygenase**-2 selective inhibitors and nonselective antiinflammatory drugs."

In this case, "cyclooxygenase" was incorrectly labelled as a drug (instead of the more accurate "cyclooxygenase-2 selective inhibitors"). The EU-ADR reports this relationship as "false" (⅔ experts agree), whereas the crowd's consensus is "positive" (80.83%). This example is further complicated by the fact that the relationship is drawn from a paper title, which are often (as in this case) only sentence fragments.

Finally, for some relations the annotations provided by the expert annotators appear to be simply incorrect. For example,

> "Exposure to Benzodiazepines (**BZD**) during foetal life has been suggested to contribute to neonatal morbidity and some congenital malformations, for example, **orofacial clefts**."

EU-ADR annotators reported this as a false relation (⅔ experts agree), whereas the crowd choice was "speculative" with an agreement score of 60.43% which we would interpret as the more correct choice.

The range in annotation consistency across experts is even more apparent when viewed at the stage prior to conflict resolution - a stage that is completely hidden from public view in most cases. Only annotations agreed upon by 2 or more out of 3 expert annotators appear in the EU-ADR; more annotations from the same text and following the same guidelines were only identified by 1 of the 3 annotators (Fig. 3). Only a small minority of the candidate annotations are supported by all the experts. In addition, not all of the annotations which received the support of two expert annotators made it into the final corpus. 51 of 298 (17.11%) annotations with two or more experts agreeing were left out, presumably in a final round of revision.

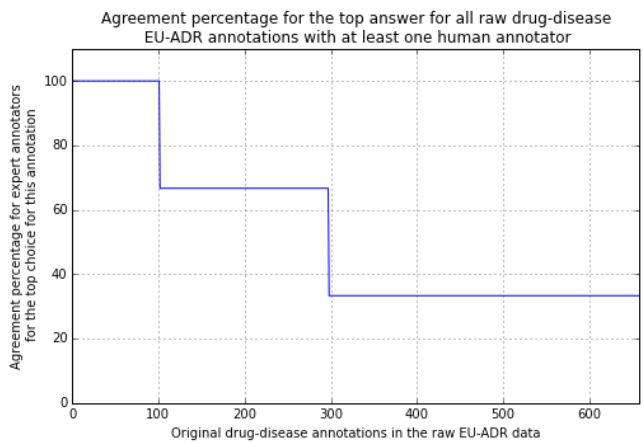

**Fig. 3.** All original drug-disease annotations in the EU-ADR and how many experts agreed upon each annotation.

## 4 DISCUSSION

This work resonates with Aroyo and Welty's "crowd truth" framework for building and applying gold standards [11]. In this work, the authors argue, as we do here, that disagreement among human annotators is not only unavoidable but is actually a source of valuable information. As an example, they show how a crowdsourcing task on medical relation extraction can generate a measure of 'sentence clarity' used to weight sentences during training and evaluation of a relation extraction system [12]. This measure corresponds directly to the crowd agreement scores presented here. Both studies bring home the simple but underappreciated truth that language can be ambiguous and that this must be factored in when building and evaluating information extraction systems. Some sentences deliver clear messages while others are mixed.

Our work reinforces the utility of crowdsourcing for gold standard creation and the general premise that disagreement between human annotators (expert and crowd) is an important aspect of such standards that should be reported. In addition to reinforcing the claims of the crowd truth research, we demonstrated how similar techniques can be used to expose the latent ambiguity and missing semantics in an existing gold standard rather than just during the creation process. Though not evaluated in depth here, we also asked our workers to define the nature of the relations indicated in the drug/disease sentences that they processed. These additional semantic data are available along with all of the data collected at (github.com/SuLab/crowdflower_relation_verification).

Beyond the ability to capture the ambiguity present in language processing tasks, crowdsourcing methods such as the one presented here offer other advantages that have yet to be understood in depth. For one, platforms such as Crowdflower and the Amazon Mechanical Turk make it possible to quickly evaluate our many different formulations of an annotation task. This makes it possible to iterate far more rapidly than would be possible without near instantaneous access to thousands of human (though not expert) annotators. Aside from rapid iteration, these systems also offer the potential to generate far larger annotated corpora than could ever before be considered. Once our task was prepared, which did take several iterations to achieve, 60 sentences were annotated by 10 different people in just 2 hours for a trivial amount of money.

This work offers a foundation to explore several future avenues of research. For example, we are exploring the use more nuanced metrics for computing crowd agreement, as well as the use of natural language processing tools that use the disagreement among crowd members in training to improve results rather than considering this factor as noise. We believe these enhancements will further increase the value and utility of crowd-based corpora. Moreover, we are working on applying crowdsourcing in concert with natural language processing and biocuration to construct a large knowledgebase of biological relationships.

## ACKNOWLEDGEMENTS

This work was supported by grants from the National Institute of General Medical Science (GM114833 and GM089820). We would also like to thank Alex Bravo Serrano and Laura Furlong from the EU-ADR for sharing the original expert annotation data.

## REFERENCES

1. Neves, M., *An analysis on the entity annotations in biological corpora.* F1000Res, 2014. **3**: p. 96.
2. Dogan, R.I., R. Leaman, and Z. Lu, *NCBI disease corpus: a resource for disease name recognition and concept normalization.* J Biomed Inform, 2014. **47**: p. 1-10.
3. Rindflesch, T.C. and M. Fiszman, *The interaction of domain knowledge and linguistic structure in natural language processing: interpreting hypernymic propositions in biomedical text.* J Biomed Inform, 2003. **36**(6): p. 462-77.
4. Ahlers, C.B., et al., *Extracting semantic predications from Medline citations for pharmacogenomics.* Pac Symp Biocomput, 2007: p. 209-20.
5. Kilicoglu, H., et al., *Constructing a semantic predication gold standard from the biomedical literature.* BMC Bioinformatics, 2011. **12**: p. 486.
6. van Mulligen, E.M., et al., *The EU-ADR corpus: annotated drugs, diseases, targets, and their relationships.* J Biomed Inform, 2012. **45**(5): p. 879-84.
7. Bravo, À., et al., *Extraction of relations between genes and diseases from text and large-scale data analysis: implications for translational research*. 2014.
8. Zhai, H., et al., *Web 2.0-based crowdsourcing for high-quality gold standard development in clinical natural language processing.* J Med Internet Res, 2013. **15**(4): p. e73.
9. Yetisgen-Yildiz, M., I. Solti, and F. Xia, *Using Amazon's Mechanical Turk for Annotating Medical Named Entities.* AMIA Annu Symp Proc, 2010. **2010**: p. 1316.
10. Good, B.M., et al., *Microtask crowdsourcing for disease mention annotation in PubMed abstracts.* Pac Symp Biocomput, 2015: p. 282-93.
11. Aroyo, L. and C. Welty, *Truth Is a Lie: Crowd Truth and the Seven Myths of Human Annotation.* AI Magazine, 2015. **36**(1): p. 15-24.
12. Aroyo, L. and C. Welty. *Measuring crowd truth for medical relation extraction*. in *2013 AAAI Fall Symposium Series*. 2013.